\documentclass[11pt]{article}

\usepackage[margin=1in]{geometry}
\usepackage{times}

\usepackage{amsmath,amsfonts,bm}

\def\eqref#1{equation~\ref{#1}}

\def\1{\bm{1}}

\DeclareMathAlphabet{\mathsfit}{\encodingdefault}{\sfdefault}{m}{sl}
\SetMathAlphabet{\mathsfit}{bold}{\encodingdefault}{\sfdefault}{bx}{n}

\usepackage[round,authoryear]{natbib}
\usepackage{hyperref}
\hypersetup{hidelinks}
\usepackage{xurl}
\usepackage{graphicx}
\usepackage{booktabs}
\usepackage{multirow}
\usepackage{amsmath}
\usepackage{amssymb}
\usepackage{float}

\title{Conformal Factuality Control for Multi-Hop Retrieval-Augmented Generation}

\author{
Muhammad Aimal Rehman\\
Department of Mathematics \& Statistics\\
Georgia State University\\
Atlanta, GA, USA\\
\texttt{mrehman3@gsu.edu}
\and
Chi-Kuang Yeh\\
Department of Mathematics \& Statistics\\
Georgia State University\\
Atlanta, GA, USA\\
\texttt{cyeh@gsu.edu}
}

\date{}

\begin{document}

\maketitle

\begin{abstract}
Retrieval-augmented generation (RAG) can ground large language models in
external evidence, but retrieved context does not guarantee that generated
claims are factually supported. This problem is especially relevant in
multi-hop RAG, where retrieval and reasoning proceed through multiple
dependent stages. We study whether claim-level conformal factuality
control, previously developed for RAG, remains effective in this setting.
We apply split-conformal claim filtering to multi-hop RAG and evaluate it
on HotpotQA, Natural Questions, and TriviaQA using Llama 3.1 8B and
GPT-4o-mini, together with a single-hop reference experiment. Across all
six multi-hop model--dataset configurations, increasingly stringent
conformal targets consistently increase the fraction of responses whose
retained claims are fully supported. At the 95\% target, this rate ranges
from 95.80\% to 97.20\%, compared with 55.60\%--76.03\% without
filtering. However, the improvement is strongly selective: only
4.41\%--31.09\% of generated claims are retained and
9.70\%--51.40\% of responses remain non-empty at the 95\% target.
These results show that conformal factuality extends to multi-hop RAG,
while demonstrating that nominal reliability must be interpreted jointly
with claim retention and abstention.
\end{abstract}

\section{Introduction}
\label{sec:introduction}
Retrieval-augmented generation (RAG) improves the ability of large
language models (LLMs) to answer knowledge-intensive questions by
conditioning generation on externally retrieved evidence
\citep{lewis2020rag}. Grounding generation in retrieved documents can
reduce dependence on parametric knowledge and provide access to
information that is more specific or up to date. Nevertheless,
retrieval alone does not guarantee that the generated response is
supported by the retrieved evidence. An LLM may still introduce
unsupported claims, combine evidence incorrectly, or generate statements
whose confidence is difficult to interpret.

This reliability problem becomes particularly important in
\emph{multi-hop} RAG. Unlike single-hop retrieval, where evidence is
typically retrieved from the original query in one retrieval stage,
multi-hop systems retrieve and reason iteratively: information obtained
at one step may determine what must be retrieved at the next
\citep{trivedi2023ircot}. Such systems are useful for compositional
questions whose answers depend on evidence distributed across multiple
documents, but they also introduce additional dependencies between
retrieval, intermediate reasoning, and final generation. Consequently,
grounding a response in retrieved evidence does not by itself provide a
principled measure of whether the claims retained in the final answer are
reliable.

Conformal prediction offers a complementary approach to this problem.
Conformal factuality uses held-out calibration data to transform a
factuality or confidence score into a statistically calibrated mechanism
for controlling the reliability of language-model outputs
\citep{mohri2024conformal}. This idea has subsequently been extended to
retrieval-augmented systems. For example, C-RAG studies conformal
generation risk for RAG models \citep{kang2024crag}, while
Conformal-RAG applies conformal prediction to sub-claims in generated
responses and uses information from retrieved documents to retain
supported content under a prescribed factuality target
\citep{feng2025conformalrag}. Recent analyses further show that such
filtering can involve a substantial trade-off between factuality and
informativeness, particularly at stringent target levels
\citep{chen2026robust}. These studies establish an important foundation
for reliable RAG, but the behavior of claim-level conformal factuality
when retrieval itself proceeds through multiple dependent hops remains
less well characterized.

In this work, we study whether conformal factuality control transfers
effectively from conventional RAG to multi-hop RAG. We apply a common
conformal filtering framework to responses generated from multi-stage
retrieval and examine not only whether increasingly stringent factuality
targets improve response reliability, but also what they cost in terms of
retained information and abstention. We first use a single-hop RAG
experiment on HotpotQA as a reference validation of the conformal
pipeline. Our primary evaluation then considers multi-hop RAG across
three question-answering benchmarks - HotpotQA, Natural Questions, and
TriviaQA - and two LLM families, Llama 3.1 8B and GPT-4o-mini. The three
benchmarks are connected to Wikipedia evidence in different ways:
HotpotQA contains Wikipedia-based multi-hop question-answer pairs; Natural
Questions uses real search queries annotated against Wikipedia pages; and
TriviaQA pairs trivia questions with evidence gathered from Wikipedia and
the broader web \citep{yang2018hotpotqa,kwiatkowski2019natural,joshi2017triviaqa}.

Our experiments reveal a consistent but non-trivial pattern. Conformal
filtering substantially increases the fraction of responses whose
retained claims are fully supported as the requested factuality level
becomes more stringent. This improvement is accompanied by progressively
lower claim retention and a higher frequency of empty responses,
demonstrating that reliability is obtained through selective output
rather than through cost-free correction of the underlying generation.
Importantly, the severity of this trade-off varies across datasets and
model families. These results suggest that conformal factuality remains
applicable in the more complex multi-hop setting, while also highlighting
that practical reliability must be evaluated jointly with the amount of
useful content preserved.

Our contributions are threefold:
\begin{itemize}
    \item We study conformal factuality control in multi-hop RAG, where
    retrieval and reasoning proceed through multiple dependent stages
    rather than a single retrieval step.

    \item We conduct a controlled evaluation across three QA datasets and
    two LLM families, together with a single-hop reference experiment,
    to examine whether the observed behavior persists across datasets and
    generation models.

    \item We characterize the reliability--informativeness trade-off in
    multi-hop RAG by jointly analyzing factual support, claim retention,
    and abstention across increasingly stringent conformal targets.
\end{itemize}

\section{Related Work}
\label{sec:related}
\paragraph{Retrieval-augmented and multi-hop generation.}
Retrieval-augmented generation combines parametric language models with
external non-parametric knowledge, enabling generated outputs to be
conditioned on retrieved evidence \citep{lewis2020rag}. While conventional
RAG commonly follows a retrieve-then-generate pattern, complex questions
may require evidence distributed across multiple documents. IRCoT
addresses this setting by interleaving retrieval with chain-of-thought
reasoning, allowing intermediate reasoning steps to guide subsequent
retrieval \citep{trivedi2023ircot}. MultiHop-RAG further demonstrates that
multi-hop queries remain challenging for both retrieval and generation,
even for strong language models \citep{tang2024multihoprag}. These findings
motivate evaluating reliability specifically when retrieval and reasoning
are sequentially dependent rather than performed in a single stage.

\paragraph{Factuality and evaluation in RAG.}
Grounding generation in retrieved documents can improve factuality, but
RAG systems still require methods for determining whether generated
answers are actually supported by their evidence. Reference-free
evaluation frameworks such as RAGAS separately assess dimensions including
faithfulness, context relevance, and answer relevance
\citep{es2024ragas}, while ARES trains task-specific lightweight judges and
uses prediction-powered inference to estimate RAG quality
\citep{saadfalcon2024ares}. Such evaluation methods are useful for
measuring RAG behavior, but they serve a different purpose from conformal
factuality: evaluation scores do not by themselves construct a calibrated
mechanism for selectively retaining claims under a prescribed reliability
level.

\paragraph{Conformal prediction for language-model reliability.}
Conformal factuality connects language-model correctness with conformal
prediction, using calibration data to obtain probabilistic guarantees while
backing off to less specific outputs when uncertainty is high
\citep{mohri2024conformal}. Within RAG, C-RAG develops conformal risk
bounds for generation risk \citep{kang2024crag}, whereas Conformal-RAG
operates at the level of generated sub-claims and incorporates retrieval
information into a relevance score used for conformal filtering
\citep{feng2025conformalrag}. Subsequent analysis has emphasized that
stronger factuality targets can substantially reduce informativeness or
produce vacuous outputs, and that calibration can be sensitive to
distribution shift \citep{chen2026robust}. Recent work on multi-turn
reasoning has also introduced MiCP, which uses conformal prediction to
allocate error budgets across reasoning turns and enable adaptive stopping
with coverage guarantees, including in adaptive RAG
\citep{zhou2026adaptive}. Our work addresses a complementary question:
rather than calibrating when a multi-turn process should terminate, we
study how claim-level conformal factuality behaves when the response itself
is produced from multi-hop retrieval, with particular attention to the
resulting trade-off among factual support, retained content, and
abstention.

\section{Background and Problem Formulation}
\label{sec:background}
\subsection{Multi-Hop Retrieval-Augmented Generation}

Let $x \in \mathcal{X}$ denote a question and let $\mathcal{K}$ denote
an external knowledge corpus. In conventional RAG, a retriever obtains a
set of documents $D$ conditioned primarily on $x$, and a language model
generates an answer using the retrieved evidence. Multi-hop RAG instead
performs a sequence of retrieval and reasoning operations. At hop
$t \in \{1,\ldots,T\}$, the system retrieves evidence
\[
D^{(t)} = \operatorname{Retrieve}
\bigl(\mathcal{K}, q^{(t)}\bigr),
\]
where the retrieval query $q^{(t)}$ may depend on the original question
and information produced during earlier hops. The accumulated evidence is
\[
D = \bigcup_{t=1}^{T} D^{(t)},
\]
and the generator produces an initial response
$\hat{y}=G(x,D)$.

This sequential structure is important for reliability. Evidence used in
the final response may originate from different retrieval stages, and an
error in an intermediate retrieval or reasoning step can affect subsequent
steps. We therefore treat multi-hop retrieval as the mechanism that
constructs the evidence available to the generator, while factuality
control is applied to the claims in the resulting response.

\subsection{Conformal Factuality for RAG}

Following Conformal-RAG \citep{feng2025conformalrag}, we decompose the
generated response into atomic sub-claims,
\[
\hat{y}=\{c_1,c_2,\ldots,c_p\}.
\]
Each claim $c$ is assigned a relevance score $R(c)$ using information from
the query and retrieved evidence. In the original Conformal-RAG scoring
function, the score is
\[
R(c)
=
\max\left(
0,\;
\max_{d \in D}
\operatorname{sim}(x,d)
\operatorname{sim}(c,d)
\right),
\]
where $\operatorname{sim}(\cdot,\cdot)$ denotes cosine similarity in the
embedding space. The score is high when a retrieved document is both
relevant to the question and semantically aligned with the claim.

For a threshold $q$, define the nested filtering operation
\[
F_q(\hat{y})
=
\{c \in \hat{y}:R(c)\ge q\}.
\]
Increasing $q$ can only remove additional claims, so that
$F_q(\hat{y}) \subseteq F_{q'}(\hat{y})$ whenever $q \ge q'$.
On a held-out calibration set, a factuality annotation function
$A(c,x,y^*,D)\in\{0,1\}$ determines whether claim $c$ is supported,
where $y^*$ denotes the reference answer. For calibration example $i$,
the conformal score is the smallest filtering threshold for which all
remaining claims are factual:
\[
S_i
=
\inf\left\{
q\ge0:
\forall q'\ge q,\;
\forall c\in F_{q'}(\hat{y}_i),\;
A(c,x_i,y_i^*,D_i)=1
\right\}.
\]

Given $n$ calibration examples and an error tolerance
$\alpha\in(0,1)$, the calibrated threshold $\hat{q}_{\alpha}$ is obtained
from the conformal quantile of $\{S_i\}_{i=1}^{n}$. At inference time,
ground-truth answers and factuality annotations are unavailable; the
system instead returns
\[
y(x;\hat{q}_{\alpha})
=
F_{\hat{q}_{\alpha}}(\hat{y}).
\]

Under the exchangeability and annotation-correctness assumptions used by
conformal factuality, calibration targets the response-level guarantee
\citep{mohri2024conformal,feng2025conformalrag}
\[
\Pr\!\left[
y^*_{\mathrm{test}}
\Rightarrow
y(x_{\mathrm{test}};\hat{q}_{\alpha})
\right]
\ge 1-\alpha,
\]
where $\Rightarrow$ denotes entailment. Thus, a target factuality of
$90\%$ corresponds to $1-\alpha=0.90$ and $\alpha=0.10$; a target of
$95\%$ corresponds to $\alpha=0.05$. Importantly, $1-\alpha$ is a
target probability for the factuality of the \emph{filtered response};
it is not the required fraction of factual claims within an individual
response.

\subsection{Problem Formulation}

Our question is whether this claim-filtering mechanism remains effective
when $\hat{y}$ is produced by multi-hop rather than single-hop retrieval.
Given the accumulated evidence from multiple retrieval stages, we apply
the same conformal principle to the generated sub-claims and study two
quantities jointly.

The first is \emph{reliability}: whether all claims retained in the final
response are supported. The second is \emph{informativeness}: how much of
the original response survives conformal filtering. Increasing the target
$1-\alpha$ generally requires a more stringent calibrated threshold and
can therefore improve reliability at the expense of retaining fewer
claims or returning an empty response. Our experiments characterize this
trade-off across retrieval settings, datasets, and language models.

\section{Method}
\label{sec:method}
\subsection{Overview}

Our pipeline augments the claim-level conformal factuality framework of
Conformal-RAG \citep{feng2025conformalrag} with an iterative multi-hop
retrieval stage. Given a question $x$, the system first performs repeated
retrieval and query refinement to construct an accumulated evidence set.
An LLM then generates an answer conditioned on this evidence. The answer
is decomposed into atomic claims, each claim is assigned a retrieval-based
relevance score, and split-conformal calibration determines the score
threshold used to retain or remove claims.

The key distinction from the original single-stage setting is therefore
the provenance of the evidence supplied to generation and scoring:
retrieved documents may originate from several dependent retrieval hops.
The conformal filtering rule itself is kept fixed, allowing us to isolate
how factuality control behaves when the upstream retrieval process becomes
multi-stage.

\begin{figure}[t]
    \centering
    \includegraphics[width=\linewidth]{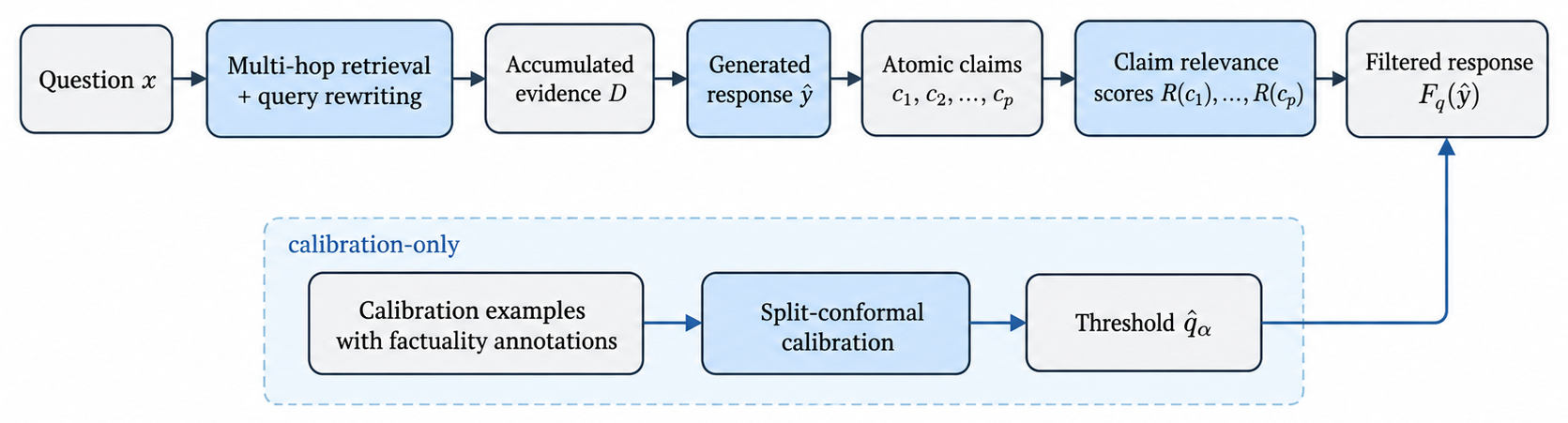}
    \caption{Overview of claim-level conformal factuality control for
    multi-hop RAG. A question triggers multi-hop retrieval with query
    rewriting and evidence accumulation. The generated response is
    decomposed into atomic claims, scored for relevance, and filtered
    using the split-conformal threshold $\hat{q}_{\alpha}$. The separate
    calibration-only branch uses calibration scores and factuality
    annotations to determine this threshold.}
    \label{fig:method-pipeline}
\end{figure}

Figure~\ref{fig:method-pipeline} summarizes the full pipeline and separates
the test-time path from the calibration-only branch, which provides the
split-conformal threshold $\hat{q}_{\alpha}$ used for final claim filtering.

\subsection{Multi-Hop Retrieval and Response Generation}

For each question, the retriever performs at most three retrieval hops.
The initial retrieval is based on the question, and subsequent retrieval
queries are produced by an LLM query rewriter using information obtained
during previous hops. At each hop, the retriever returns up to $k=10$
documents subject to a similarity threshold of $0.3$. Retrieved documents
are accumulated across hops, and retrieval terminates early when a hop
returns no previously unseen documents.

For query rewriting, at most five retrieved documents are supplied as
context, with each document truncated to 1200 characters. The query
rewriter is run deterministically with temperature zero. After the final
hop, the accumulated evidence set $D$ is provided to the response
generator to produce
\[
\hat{y}=G(x,D).
\]
This design allows evidence discovered at an earlier stage to influence
later retrieval while preserving all retrieved evidence for final answer
generation.

\subsection{Atomic Claim Decomposition and Verification}

The generated answer is decomposed into atomic sub-claims
\[
\hat{y}=\{c_1,\ldots,c_p\}.
\]
The decomposition model is instructed to express the answer as
independently verifiable factual statements. To make generation robust to
malformed structured outputs, decomposition is retried up to three times
when the output cannot be parsed into a valid set of aligned atomic
claims.

For calibration and evaluation, each claim is assigned a factuality
annotation by an LLM verifier. The verifier receives the original
question, the reference answer, the retrieved evidence, and the claim,
and returns one of four labels:
\emph{supported} ($S$), \emph{irrelevant} ($I$),
\emph{unverifiable} ($U$), or \emph{non-factual} ($N$).
Only $S$ is treated as a supported claim. These annotations are used for
calibration and evaluation; they are not required for filtering a new
test-time response once the conformal threshold has been calibrated.

\subsection{Retrieval-Based Claim Scoring}
\label{sec:claim-scoring}

For each claim $c$, we use the relevance score introduced in
Conformal-RAG:
\[
R(c)
=
\max_{d\in D}
\left[
\operatorname{sim}(x,d)
\operatorname{sim}(c,d)
\right]_+ ,
\]
where $[\cdot]_+=\max(0,\cdot)$ and cosine similarities are computed from
sentence embeddings. We use maximum aggregation across retrieved
documents and the product scoring rule. A small deterministic
tie-breaking perturbation $\epsilon(c)$ is added to break score ties, yielding
\[
\widetilde{R}(c)=R(c)+\epsilon(c).
\]

For a calibrated threshold $\hat{q}_{\alpha}$, the retained response is
therefore
\[
F_{\hat{q}_{\alpha}}(\hat{y})
=
\left\{
c\in\hat{y}:
\widetilde{R}(c)\ge\hat{q}_{\alpha}
\right\}.
\]
Importantly, documents retrieved at all multi-hop stages contribute to
the accumulated evidence available when claim relevance is scored.

\subsection{Split-Conformal Calibration}

We use split conformal calibration with factuality requirement $a=1$,
meaning that a successful filtered response contains no unsupported
retained claim. For each experimental run, the available examples are
divided deterministically into equally sized calibration and test sets.
For the 60-question experiments, this gives 30 calibration examples and
30 test examples. We repeat this procedure over 100 deterministic seeded
calibration--test splits and report averages across splits.

For a target factuality level $1-\alpha$, the calibration examples
determine the finite-sample conformal threshold
$\hat{q}_{\alpha}$. We evaluate targets ranging from $80\%$ to $95\%$,
with particular emphasis on the $90\%$ ($\alpha=0.10$) and $95\%$
($\alpha=0.05$) conditions. Within a model--dataset configuration, the
same generated responses and claim scores are reused across target levels;
only the calibrated filtering threshold changes. This isolates the effect
of conformal selectivity from variation in the underlying generation.

\subsection{Reliability and Selectivity}

A response is considered \emph{fully supported} when every retained claim
is labeled $S$. Following the conformal formulation, an empty retained
set satisfies this event vacuously. Consequently, response-level
factuality alone can obscure an important failure mode: a stringent
threshold may achieve high reliability simply by removing nearly all
content.

We therefore evaluate conformal filtering jointly along three dimensions:
(i) factual support of retained claims, (ii) the fraction of generated
claims retained, and (iii) response coverage, measured by the fraction of
answers that retain at least one claim. We additionally report the
fully-supported rate conditional on a response being non-empty. These
metrics distinguish genuinely useful reliable answers from reliability
obtained primarily through abstention.

\section{Experimental Setup}
\label{sec:experiments}

\subsection{Datasets}

We evaluate HotpotQA \citep{yang2018hotpotqa}, Natural Questions (NQ)
\citep{kwiatkowski2019natural}, and TriviaQA
\citep{joshi2017triviaqa}. HotpotQA explicitly targets multi-document
reasoning, while NQ and TriviaQA provide complementary open-domain QA
settings.

We use a controlled 60-question evaluation subset for each benchmark.
The HotpotQA subset is sampled from the KILT validation version
\citep{petroni2021kilt} with seed 42 after requiring a non-empty answer,
at least two distinct provenance documents, and compatibility with the
Wikipedia retrieval corpus. NQ and TriviaQA are likewise evaluated on
fixed 60-question subsets. Using the same sample size across benchmarks
keeps generation, verification, and repeated conformal calibration
computationally controlled.

\subsection{Models and Experimental Conditions}

We evaluate two substantially different generation backbones:
\path{meta-llama/Llama-3.1-8B-Instruct}
\citep{grattafiori2024llama} and \path{gpt-4o-mini}
\citep{openai2024gpt4omini}. For the GPT-4o-mini experiments, we pin
the model snapshot to \path{gpt-4o-mini-2024-07-18} for
reproducibility. The Llama configuration uses
\path{meta-llama/Llama-3.1-8B-Instruct} for response generation,
query rewriting, atomic-claim decomposition, and claim verification.
The pinned GPT-4o-mini snapshot is used for the corresponding four
model-driven stages in the GPT-4o-mini experiments. Sentence embeddings
for both model families are produced with
\path{sentence-transformers/all-MiniLM-L6-v2}.

Our primary experiments use multi-hop retrieval for both model families
on all three datasets. In addition, we evaluate Plain RAG with
Llama 3.1 8B on HotpotQA as a reference condition for confirming the
behavior of the conformal pipeline in the simpler single-hop setting.
Table~\ref{tab:experimental-matrix} summarizes the evaluation matrix.

\begin{table}[t]
\centering
\caption{Experimental configurations. The Plain-RAG setting serves as a
reference experiment; the six Multi-Hop RAG configurations constitute the
primary evaluation.}
\label{tab:experimental-matrix}
\small
\begin{tabular}{lll}
\toprule
\textbf{Retrieval} & \textbf{Model} & \textbf{Datasets} \\
\midrule
Plain RAG &
Llama 3.1 8B &
HotpotQA \\
Multi-Hop RAG &
Llama 3.1 8B &
HotpotQA, NQ, TriviaQA \\
Multi-Hop RAG &
GPT-4o-mini &
HotpotQA, NQ, TriviaQA \\
\bottomrule
\end{tabular}
\end{table}

\subsection{Conformal Evaluation Protocol}

For each model--dataset configuration, generation and claim scoring are
performed once, and the resulting examples are reused across conformal
targets. We use split conformal calibration with $a=1$. Each 60-example
set is divided into 30 calibration and 30 test examples, and results are
averaged over 100 seeded calibration--test splits.

We evaluate
\[
1-\alpha \in \{0.80,0.85,0.90,0.925,0.95\},
\]
with primary emphasis on the $90\%$ ($\alpha=0.10$) and $95\%$
($\alpha=0.05$) targets. The unfiltered condition retains all generated
claims. The 100 runs are repeated partitions of fixed generated examples,
not independent end-to-end generations.

\section{Results}
\label{sec:results}

\subsection{Reference Single-Hop Experiment}

On HotpotQA with Llama 3.1 8B, the Plain-RAG baseline yields
$61.67\%$ fully supported responses. At the $90\%$ target, this rises
to $89.73\%$, with $41.25\%$ claim retention and $61.07\%$ non-empty
responses. At the $95\%$ target, full support reaches $96.27\%$, while
claim retention falls to $11.10\%$ and non-empty responses to $20.97\%$.
This reference condition therefore exhibits the expected
reliability--selectivity trade-off.

\subsection{Conformal Filtering Raises Response-Level Support in Multi-Hop RAG}

Table~\ref{tab:main-results} reports the primary multi-hop results.
Without filtering, fully supported responses range from $55.60\%$ to
$76.03\%$ across the six configurations. At the $90\%$ target,
empirical full-support rates range from $88.70\%$ to $91.33\%$; at
$95\%$, they range from $95.80\%$ to $97.20\%$. The same qualitative
pattern appears for both Llama 3.1 8B and GPT-4o-mini across all three
datasets.

\begin{table}[t]
\centering
\caption{
Main Multi-Hop RAG results. FS denotes the percentage of responses
that are fully supported under the response-level conformal event,
NE denotes non-empty responses, and Ret denotes generated claims
retained after conformal filtering. All values are percentages and
are averaged over 100 calibration--test splits.
}
\label{tab:main-results}
\small
\setlength{\tabcolsep}{3.5pt}
\begin{tabular}{@{}llr|rrr|rrr@{}}
\toprule
& &
\multicolumn{1}{c|}{\textbf{Base}} &
\multicolumn{3}{c|}{\textbf{90\% target}} &
\multicolumn{3}{c}{\textbf{95\% target}} \\
\textbf{Model} &
\textbf{Dataset} &
\textbf{FS} &
\textbf{FS} &
\textbf{NE} &
\textbf{Ret} &
\textbf{FS} &
\textbf{NE} &
\textbf{Ret} \\
\midrule
Llama 3.1 8B
& HotpotQA
& 68.60 & 91.13 & 60.57 & 32.39
& 96.90 & 22.00 & 8.35 \\

& NQ
& 76.03 & 88.70 & 66.53 & 41.10
& 95.80 & 31.90 & 14.94 \\

& TriviaQA
& 55.60 & 89.93 & 63.03 & 40.52
& 96.50 & 51.40 & 28.17 \\
\midrule
GPT-4o-mini
& HotpotQA
& 64.80 & 90.30 & 51.80 & 27.86
& 97.20 & 10.80 & 5.16 \\

& NQ
& 57.70 & 90.23 & 31.97 & 19.73
& 96.43 & 9.70 & 4.41 \\

& TriviaQA
& 70.60 & 91.33 & 78.03 & 60.31
& 96.37 & 48.43 & 31.09 \\
\bottomrule
\end{tabular}
\end{table}

\subsection{Reliability Comes at a Substantial Selectivity Cost}

Higher response-level factuality is accompanied by substantial loss of
retained information. At the $90\%$ target, claim retention ranges from
$19.73\%$ to $60.31\%$ and non-empty response coverage from $31.97\%$
to $78.03\%$. At the $95\%$ target, these ranges fall to
$4.41\%$--$31.09\%$ and $9.70\%$--$51.40\%$, respectively.

\begin{figure}[t]
    \centering
    \includegraphics[width=\linewidth]
    {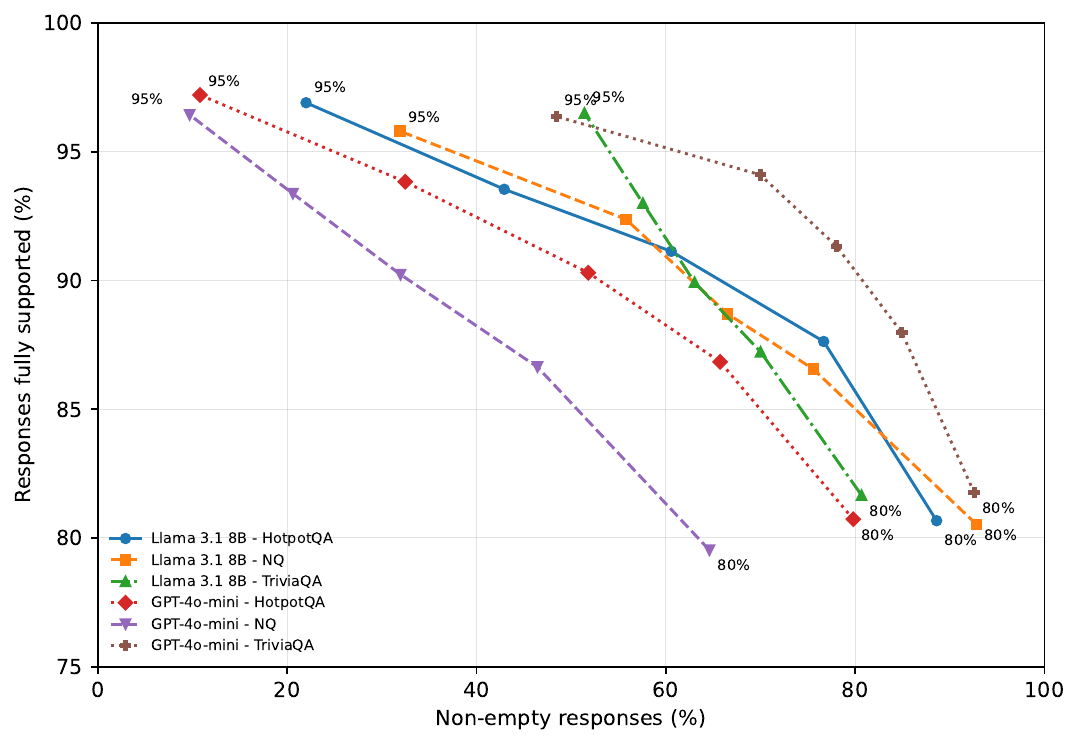}
    \caption{
    Reliability--coverage trade-off for Multi-Hop RAG as the conformal
    target increases from $80\%$ to $95\%$. Each trajectory corresponds
    to one model--dataset configuration; endpoint labels indicate the
    $80\%$ and $95\%$ targets. More stringent conformal filtering
    increases the fraction of responses satisfying the full-support event
    while generally decreasing the fraction of responses that retain any
    content.
    }
    \label{fig:reliability-coverage}
\end{figure}

Figure~\ref{fig:reliability-coverage} shows that this cost varies strongly
by dataset. At the $95\%$ target, TriviaQA retains substantially more
content than GPT-4o-mini on HotpotQA and NQ, where claim retention falls
to $5.16\%$ and $4.41\%$, respectively.

\subsection{Response-Level Reliability Can Be Dominated by Abstention}

Because empty filtered responses satisfy the full-support event
vacuously, high response-level factuality need not imply equally high
reliability among non-empty answers. At the $95\%$ target,
GPT-4o-mini on HotpotQA and NQ achieves overall full-support rates of
$97.20\%$ and $96.43\%$, while only $10.80\%$ and $9.70\%$ of responses
remain non-empty. Conditional on non-empty output, the corresponding
fully-supported rates are only $78.48\%$ and $71.72\%$.

These results show why response-level factuality should be interpreted
jointly with claim retention and non-empty response coverage.

\section{Discussion}
\label{sec:discussion}

Across all six multi-hop model--dataset configurations, increasingly
stringent conformal targets raise the response-level rate at which all
retained claims are supported. This pattern holds for both Llama 3.1 8B
and GPT-4o-mini and across HotpotQA, NQ, and TriviaQA, indicating that
claim-level conformal factuality remains applicable when the upstream
retrieval process is multi-hop.

The improvement, however, should be interpreted as \emph{selective
factuality control} rather than correction of the underlying generation.
Higher targets remove increasingly more generated content and can produce
empty responses. Because an empty filtered response satisfies the
response-level full-support event vacuously, high overall factuality can
coexist with low non-empty response coverage. This is especially visible
for GPT-4o-mini on HotpotQA and NQ at the $95\%$ target. Our results
therefore reinforce the need to report conformal factuality jointly with
claim retention and abstention \citep{chen2026robust}.

The operational cost of a given target also varies substantially across
model--dataset combinations. TriviaQA retains considerably more content
at stringent targets than several HotpotQA and NQ settings, showing that
the same nominal reliability target need not imply the same practical
utility. Target selection should therefore depend on application-specific
tolerance for abstention and information loss. Finally, the Plain-RAG
experiment serves only as a reference validation on HotpotQA with
Llama 3.1 8B; it does not support a general comparison between single-hop
and multi-hop RAG.

\section{Limitations}
\label{sec:limitations}

Our study has several limitations. First, the experiments use controlled
60-question subsets rather than complete benchmark evaluation sets. This
makes repeated generation, verification, and calibration tractable, but
limits the precision with which the reported percentages estimate
full-benchmark performance.

Second, the 100 conformal runs are repeated calibration--test partitions
of fixed generated responses, not independent end-to-end generations.
The reported variation therefore reflects calibration-sample sensitivity
rather than stochastic variability in retrieval or generation.

Third, factuality annotations rely on an LLM-based verifier and thus
inherit possible annotation errors and model-dependent judgments about
support. The conformal interpretation also depends on exchangeability
between calibration and test examples and on annotation quality, so
distribution shift or systematic verifier errors can weaken the practical
meaning of the nominal target.

Fourth, atomic claim decomposition is automated and model-dependent.
Although malformed structured outputs are retried, decomposition errors
can still affect downstream scoring, verification, and response-level
metrics.

Finally, the Plain-RAG experiment is only a reference condition on
HotpotQA with Llama 3.1 8B, so we do not draw a general conclusion about
whether conformal factuality is more or less costly in single-hop versus
multi-hop RAG. Our evaluation also covers only two model families and one
multi-hop retrieval design with at most three hops; other models,
retrieval strategies, domains, and longer reasoning chains may exhibit
different trade-offs.

\section{Conclusion}
\label{sec:conclusion}

We studied claim-level conformal factuality control for multi-hop RAG
across HotpotQA, Natural Questions, and TriviaQA using Llama 3.1 8B and
GPT-4o-mini. Across all six multi-hop configurations, more stringent
conformal targets increased the response-level full-support rate, but did
so by retaining fewer claims and, in some settings, producing many empty
responses.

These results show that conformal factuality can be applied to multi-hop
RAG, while also demonstrating that reliability targets should be
interpreted jointly with claim retention and response coverage. The
practical value of conformal filtering therefore depends not only on the
nominal target, but also on the amount of useful content preserved.

\subsection*{Reproducibility statement}

We provide the methodological and experimental details required to
reproduce the reported study throughout the paper. The implementation
used in this work is publicly available at
\url{https://github.com/aimalrehman92/conformal-rag}.
Section~\ref{sec:method} specifies the multi-hop retrieval, claim
decomposition, verification, scoring, split-conformal calibration, and
evaluation metrics, while Section~\ref{sec:experiments} documents the
evaluated datasets, language models, conformal targets, calibration
protocol, and random seed. Additional implementation and experimental
details, including the complete conformal target sweeps and configuration
information, are provided in the appendix. The reported results are
computed from fixed generated-response artifacts and repeated seeded
calibration--test splits, as described in the experimental protocol.

\bibliography{iclr2027_conference}
\bibliographystyle{plainnat}

\appendix

\section{Additional Experimental Results}
\label{app:additional-results}
The main paper reports the unfiltered baseline together with the
$90\%$ and $95\%$ conformal conditions. Here we provide the complete
target sweeps used to characterize the reliability--informativeness
trade-off. In addition to claim retention and non-empty response
coverage, we report supported-claim factuality and the fraction of
non-empty responses for which every retained claim is supported.
The latter excludes vacuous empty responses and therefore complements
the response-level conformal event reported in the main text.

\subsection{Plain-RAG Reference Sweep}

\begin{table}[H]
\centering
\caption{Complete conformal target sweep for the Plain-RAG reference
experiment on HotpotQA with Llama 3.1 8B. All values are percentages.}
\label{tab:appendix-plain-sweep}
\small
\begin{tabular}{rrrrr}
\toprule
\textbf{Target} &
\textbf{Non-empty} &
\textbf{Claims retained} &
\textbf{Supported claims} &
\textbf{Fully supported $\mid$ non-empty} \\
\midrule
80.0 & 81.43 & 67.40 & 90.34 & 76.61 \\
85.0 & 72.20 & 52.96 & 91.32 & 82.87 \\
90.0 & 61.07 & 41.25 & 91.01 & 85.16 \\
92.5 & 44.83 & 25.82 & 88.99 & 86.98 \\
95.0 & 20.97 & 11.10 & 86.67 & 88.51 \\
\bottomrule
\end{tabular}
\end{table}

\subsection{Multi-Hop RAG with Llama 3.1 8B}

\begin{table}[H]
\centering
\caption{Complete conformal target sweeps for Multi-Hop RAG with
Llama 3.1 8B. ``Supported claims'' is supported-claim factuality;
the final column is response-level full support conditional on the
filtered response being non-empty. All values are percentages.}
\label{tab:appendix-llama-sweep}
\small
\setlength{\tabcolsep}{5pt}
\begin{tabular}{llrrrr}
\toprule
\textbf{Dataset} &
\textbf{Target} &
\textbf{Non-empty} &
\textbf{Claims retained} &
\textbf{Supported claims} &
\textbf{Fully supported $\mid$ non-empty} \\
\midrule
HotpotQA & 80.0 & 88.60 & 68.39 & 88.22 & 78.65 \\
         & 85.0 & 76.67 & 47.01 & 90.33 & 84.95 \\
         & 90.0 & 60.57 & 32.39 & 90.47 & 87.06 \\
         & 92.5 & 42.93 & 20.16 & 89.67 & 86.83 \\
         & 95.0 & 22.00 &  8.35 & 93.54 & 90.83 \\
\midrule
NQ       & 80.0 & 92.80 & 82.99 & 92.60 & 79.75 \\
         & 85.0 & 75.57 & 53.31 & 93.61 & 83.62 \\
         & 90.0 & 66.53 & 41.10 & 93.90 & 84.48 \\
         & 92.5 & 55.80 & 29.57 & 95.21 & 88.53 \\
         & 95.0 & 31.90 & 14.94 & 96.62 & 92.63 \\
\midrule
TriviaQA & 80.0 & 80.67 & 62.60 & 86.53 & 78.15 \\
         & 85.0 & 70.00 & 48.27 & 89.87 & 83.20 \\
         & 90.0 & 63.03 & 40.52 & 92.08 & 85.44 \\
         & 92.5 & 57.57 & 34.38 & 94.48 & 88.61 \\
         & 95.0 & 51.40 & 28.17 & 97.07 & 93.78 \\
\bottomrule
\end{tabular}
\end{table}

\subsection{Multi-Hop RAG with GPT-4o-mini}

\begin{table}[H]
\centering
\caption{Complete conformal target sweeps for Multi-Hop RAG with
GPT-4o-mini. All values are percentages.}
\label{tab:appendix-openai-sweep}
\small
\setlength{\tabcolsep}{5pt}
\begin{tabular}{llrrrr}
\toprule
\textbf{Dataset} &
\textbf{Target} &
\textbf{Non-empty} &
\textbf{Claims retained} &
\textbf{Supported claims} &
\textbf{Fully supported $\mid$ non-empty} \\
\midrule
HotpotQA & 80.0 & 79.80 & 53.44 & 88.96 & 76.91 \\
         & 85.0 & 65.73 & 38.19 & 90.46 & 81.61 \\
         & 90.0 & 51.80 & 27.86 & 90.67 & 83.82 \\
         & 92.5 & 32.47 & 16.14 & 90.36 & 85.07 \\
         & 95.0 & 10.80 &  5.16 & 85.48 & 78.48 \\
\midrule
NQ       & 80.0 & 64.60 & 44.69 & 87.12 & 70.28 \\
         & 85.0 & 46.43 & 29.56 & 88.60 & 74.34 \\
         & 90.0 & 31.97 & 19.73 & 87.99 & 73.89 \\
         & 92.5 & 20.60 & 11.97 & 86.56 & 73.48 \\
         & 95.0 &  9.70 &  4.41 & 82.10 & 71.72 \\
\midrule
TriviaQA & 80.0 & 92.57 & 81.43 & 89.62 & 80.71 \\
         & 85.0 & 84.93 & 68.95 & 92.40 & 86.37 \\
         & 90.0 & 78.03 & 60.31 & 93.66 & 89.51 \\
         & 92.5 & 69.97 & 50.14 & 95.27 & 92.43 \\
         & 95.0 & 48.43 & 31.09 & 96.88 & 95.05 \\
\bottomrule
\end{tabular}
\end{table}

\section{Implementation Details}
\label{app:implementation}
\subsection{Retrieval and Indexing Configuration}

For the Llama 3.1 8B experiments, documents are embedded with
\path{sentence-transformers/all-MiniLM-L6-v2} and stored in a FAISS
index. Documents are segmented using fixed-length chunks of 2000
characters with an overlap of 25 characters. Multi-hop retrieval returns
up to $k=10$ documents per hop using a similarity threshold of $0.3$.
The system performs at most three retrieval hops, accumulates documents
across hops, and terminates early when a hop retrieves no previously
unseen document.

Query rewriting is performed deterministically with temperature zero.
The rewriter conditions on at most five retrieved documents, each
truncated to 1200 characters. For the Llama 3.1 8B configuration,
\path{meta-llama/Llama-3.1-8B-Instruct} is used for answer generation,
query rewriting, atomic-claim decomposition, and claim verification.
Answer generation is also performed with temperature zero.
For the GPT-4o-mini experiments, the OpenAI model snapshot
\path{gpt-4o-mini-2024-07-18} is pinned for reproducibility. The same
snapshot is used for response generation, multi-hop query rewriting,
atomic-claim decomposition, and claim verification.
The local embedding model and retrieval configuration are kept aligned
with the Llama 3.1 8B experiments: embeddings use
\path{sentence-transformers/all-MiniLM-L6-v2}, retrieval uses
$k=10$ with a similarity threshold of $0.3$ and at most three hops,
retrieved documents are accumulated across hops, and retrieval stops
early when no previously unseen document is returned. The GPT-4o-mini
generator and query rewriter use temperature zero; the query rewriter
conditions on at most five documents truncated to 1200 characters each.

\subsection{Claim Processing}

Generated responses are decomposed into independently verifiable atomic
claims. Because structured model outputs can occasionally be malformed, the implementation retries atomic decomposition up to
three times when the generated output cannot be parsed into a non-empty
set of claims aligned with the corresponding token-probability groups.

Claim verification likewise uses a structured-output procedure with up to
three attempts when the verifier response cannot be parsed into a valid
label. Each claim receives one of four annotations:
\emph{supported} ($S$), \emph{irrelevant} ($I$),
\emph{unverifiable} ($U$), or \emph{non-factual} ($N$).
Only $S$ is treated as factual when computing the conformal event and
supported-claim factuality.

\subsection{Conformal Configuration}

The implementation uses split conformal calibration with maximum
aggregation across retrieved documents and the product relevance-scoring
rule described in Section~\ref{sec:claim-scoring}. The factuality parameter
is fixed to $a=1$, requiring every retained claim to be supported for the
response-level event to hold. We use a fixed random seed of 42 and 100
repeated calibration--test splits.

The headline conditions correspond to
$\alpha=0.10$ and $\alpha=0.05$, i.e., target factuality levels of
$90\%$ and $95\%$. For the extended trade-off analysis, we additionally
evaluate target levels of $80\%$, $85\%$, and $92.5\%$. Within each
model--dataset configuration, the same generated responses, claim scores,
and verifier annotations are reused across conformal targets so that only
the calibrated threshold changes.

\subsection{Execution Environment}

The representative Llama 3.1 8B HotpotQA experiment was executed in Google
Colab with Python 3.13.15, PyTorch 2.11.0 with CUDA 12.8 support, and an
NVIDIA A100-SXM4-40GB GPU. The run used 60 controlled HotpotQA questions,
seed 42, and 100 conformal evaluation splits. The local embedding model
was \path{sentence-transformers/all-MiniLM-L6-v2}.

\end{document}